# Image Captioning with Attention for Smart Local Tourism using EfficientNet

**Dhomas Hatta Fudholi, Yurio Windiatmoko, Nurdi Afrianto, Prastyo Eko Susanto, Magfirah Suyuti, Ahmad Fathan Hidayatullah, Ridho Rahmadi**

hatta.fudholi@uii.ac.id, yurio.windiatmoko@students.uii.ac.id,
nurdi.afrianto@students.uii.ac.id, prastyo.susanto@students.uii.ac.id,
maghfirah.suyuti@students.uii.ac.id, fathan@uii.ac.id, ridho.rahmadi@uii.ac.id

Department of Informatics, Universitas Islam Indonesia, Yogyakarta, Indonesia

**Abstract**. Smart systems have been massively developed to help humans in various tasks. Deep Learning technologies push even further in creating accurate assistant systems due to the explosion of data lakes. One of the smart system tasks is to disseminate 'users needed information'. This is crucial in the tourism sector to promote local tourism destinations. In this research, we design a model of local tourism specific image captioning, which later will support the development of AI-powered systems that assist various users. The model is developed using a visual Attention mechanism and uses the state-of-the-art feature extractor architecture EfficientNet. A local tourism dataset is collected and is used in the research, along with two different kinds of captions. Captions that describe the image literally and captions that represent human logical responses when seeing the image. This is done to make the captioning model more humane when implemented in the assistance system. We compared the performance of two different models using EfficientNet architectures (B0 and B4) with other well known VGG16 and InceptionV3. The best BLEU scores we get are 73.39 and 24.51 for the training set and the validation set respectively, using EfficientNetB0. The captioning result using the developed model shows that the model can produce logical caption for local tourism-related images.

## 1. Introduction
Tourism has a major influence on economic development. The increase in population creates huge potential in tourism and has an impact on tourism development [1]. This includes the rapid growth in tourism villages. Tourists are becoming more interested in natural and cultural experiences [24]. Yogyakarta is one of the main tourist destinations in Indonesia which improves the economy around it's tourism destination [2]. Tourists are an asset [3] to develop regional economics. Rural tourism destinations in Yogyakarta may face difficulties competing with other popular tours due to a lack of information [4]. Also, various forms of service facilities (lodging, equipment, etc.) and information should be maintained [24]. Therefore, one of the Indonesian government's programs in the tourism sector is to reduce social inequality between rural and urban areas (Gini Index up to 0.4) through community-based rural development programs [5].

Artificial Intelligence (AI) and the emergence of Deep Learning technologies have become the core of smart systems in various domains, which help people in many ways. One implementation of AI-powered systems is a chatbot. Chatbots are a very valuable technology that can help in collecting and managing information related to the services of the company or a corporation [6]. Chatbots provide a new perspective on doctor-patient (DP) and clinic-patient (KP) services in terms of cost and time [7]. Chatbots handle problems related to getting customer service requests quickly and accurately [8]. Chatbots can operate 24/7 (24 hours and 7 days) in helping retailers to improve customer service and increase efficiency [9]. A chatbot can be used for patients undergoing chemotherapy remotely, of course with the supervision of skilled nurses [10]. Chatbots are used in the university environment to answer student FAQs interactively, by using the AIML and LSA approaches as well [10]. The ability of chatbot that changes the interaction between customers and organizations in all sectors involved [25] may be one of the solutions in providing tourism information ubiquitously.

The interaction within chatbot systems may be pushed further to not only text-based communication. In the tourism industry, it is really interesting to have fluid conversations based on photos as well. Image captioning technology comes to enable this feature. Image captioning uses human vision to collect information in a visual scene. Humans tend to focus on important and interesting objects in the visual scene. One of the works in image captioning using the SILICON dataset [11]. Besides, describing an image may be a difficult job because it not only recognizes objects in an image but also converts visual understanding into sentences [12]. However, giving information through photos can attract tourists to their favorite destination.

In this research, we focus on developing a model of image captioning which is specialized in captioning local tourism-related photos. We use a local tourism-related dataset from tourism destinations in Yogyakarta, Indonesia. To develop the model, we use a Deep Learning technique that uses state-of-the-art feature extraction architecture EfficientNet in the encoder side and GRU in the decoder side. The visual Attention mechanism is also applied to model architecture. Furthermore, we compare different architectures and measure their performance as the evaluation.

## 2. Literature Review

Image captioning aims to set the image description automatically in a good term language [13]. An interesting thing comes when it does not only read the image based on its visual content but also from the linguistic context of the image [14]. Some of the benefits of image captioning is to help create a description of an image for blind people in seeing the meaning of images [19]. The ability to translate visual content into natural language can be useful in various fields, such as human interaction with robots, real-time descriptions of car scenes for the visually impaired, and instantaneous display of traffic conditions to the public [20]. The image description creation model does not only capture objects in the image but also reveals the relationship between objects into a sentence [27]. Besides being useful for blind people, image captioning is also useful for searching images using text [16]. The description of an image in the industrial world can be used to share photos on social media and interact with robots [21]. Combined with smart wearable devices, image captioning can help to create personal notes based on photos collected from supporting devices [17]. Image captioning is a multidisciplinary knowledge of computer vision and Natural Language Processing that is attractive to academics and the industrial world [26].

Deep Learning techniques become the core in building an image captioning model. A study in [12] focuses on captioning images using bidirectional LSTM. In the study, augmentation techniques such as multi-crop and multi-scale were used to prevent overfitting. While other research tends to use semantic model attention to generate image descriptions, such as using the top-down and bottom-up strategy to extract more image information which has good results because it observes all the information contained in the image [19]. Another approach in image captioning is to cluster the area of attention with structural proximity. This approach models the dependencies between drawing image, word's description, and language model by RNN [27]. Another study of image captioning which takes other languages besides English is [23]. This study presented two models of Nepali-based models. The first model comprises an encoder (ResNet-50) and a decoder (a plain LSTM). The second model comprises an encoder (InceptionV3) and a decoder (GRU network with attention). However, the result of this paper shows perplexity in comparing model A and model B, the results on the test samples depend on the priors seen during the training regime. The lack of datasets and the fact that Nepali translation is not very accurate to make this research could be improved for both the models with the further tuning of the architectures and training.

To enhance the contextual aspect in natural language sequence, an Attention mechanism can be applied. The description of the image content with attention is compatible with human intuition [28]. The accuracy of attention to an image using the evaluation matrix has a positive correlation. Even so, it needs to be improved to what extent the accuracy of attention is consistent with human perceptions [15]. The model of captioning with attention area consists of three parts: combining image areas, word captions, and the NLP natural language model (RNN). Usually, the MS COCO dataset is used for testing the trained model [16]. Other research shows the AoA (Attention on Attention) method for encoder and decoder in image captioning with a score of 129.8 (CIDEr-D) on the COCO dataset [18].

Generating a caption that sensibly describes an image is key to scene understanding in artificial intelligence [23]. Extracting coherent features of an image using an image-based model is one of the processes in image captioning. An image is described into the natural language from the extracted features of the image-based model. We propose to develop a model of image captioning in the domain of local tourism to support better engagement in smart tourism systems. We use state-of-the-art architecture EfficientNet for the feature extraction process and GRU architecture in caption generating decoder. The Attention layer is applied to get a more sensible caption of the image.

## 3. Method

The local tourism becomes our domain in the research. Hence, we firstly collect a dataset of local tourism-related images and write captions for them. Secondly, we develop an image captioning model using EfficientNet as the base architecture. Finally, we test and evaluate the model using the BLEU score and compare the EfficientNet-based model with other models developed using well-known architecture. Each research stage is elaborated in detail below.

*3.1 Data Collection and Caption Annotation*

A total of 1696 local tourism-related images are collected through Google search engines using various keywords that are related to the local tourism destination in Yogyakarta, Indonesia. These keywords are 'monumen tugu' (Tugu monumen) , 'arung jeram' (rafting), 'malioboro' (Malioboro), 'pendaki wanita' (female hikers), 'wisata jeep kaliadem' (Kaliadem jeep tours), 'merapi park' (Park Merapi), 'air terjun sri gethuk' (Sri Gethuk waterfall), 'parangtritis' (Parangtritis), 'hutan pinus' (pine forest), 'bukit paralayang' (Paralayang hill), 'heha sky view' (Heha Sky View), 'ayunan langit watu jaran' (Watu Jaran sky swing), 'ice cream world', 'grojogan watu purbo' (Grojogan Watu Purbo), 'bukit isis' (Isis hill), 'bukit mojo gumelem' (Mojo Gumelem hill) and 'puncak segoro' (Segoro peak). All these keywords are closely related to tourism destinations in the local area of Yogyakarta, Indonesia.

The collected images have already been filtered by their quality to an only selected high resolution that is considered as a sharp image. The next step after collecting images is making a unique name for each image in a folder by running a randomized filename script. Finally, we made a list of annotations in .txt format and created a cross-checked list that maps images with their captions as in Figure 1.

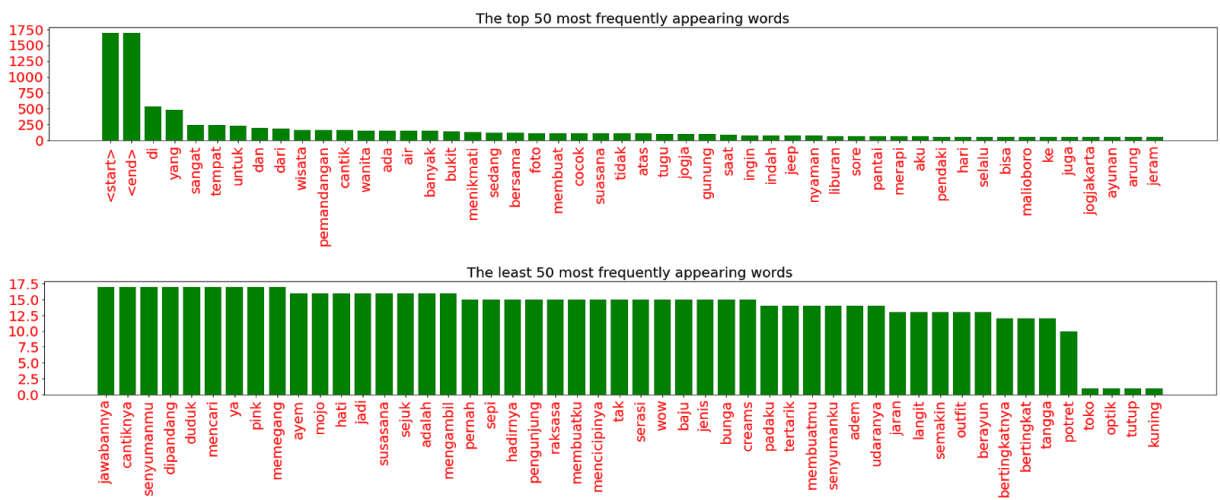

**Figure 1.** Cross-check files of images and their captions.

Speaking of the captions in the datasets. The vocabulary size of the captions is 291. Figure 2 shows the top 50 and the least 50 most frequent words in the vocabulary. From the figure, we can see that there are conjunctions that we kept, such as *'yang'* (that) and *'dan'* (and). The image captioning task requires that the output caption forms complete logical sentences. It is different from other text analytics tasks, such as sentiment analysis where those words may be omitted through stopword removal tasks since it is not necessary and becomes noises.

The dataset is divided into a training set and a validation set with an 80:20 ratio. We get 1356 data within the training set and 340 data for the validation set, which are transformed into tensor during the training through the Keras Tensorflow pipeline.

**Figure 2.** The top and the least 50 most frequently words

*3.2. Modeling*

Encoder-decoder architecture is implemented during the training process. As for the encoder, we applied pre-trained ImageNet weight transfer learning from two different EfficientNet architectures, namely EfficientNetB0 and EfficientNetB4, and two comparator architectures, namely VGG16 and InceptionV3. Figure 3 illustrates our image captioning network.

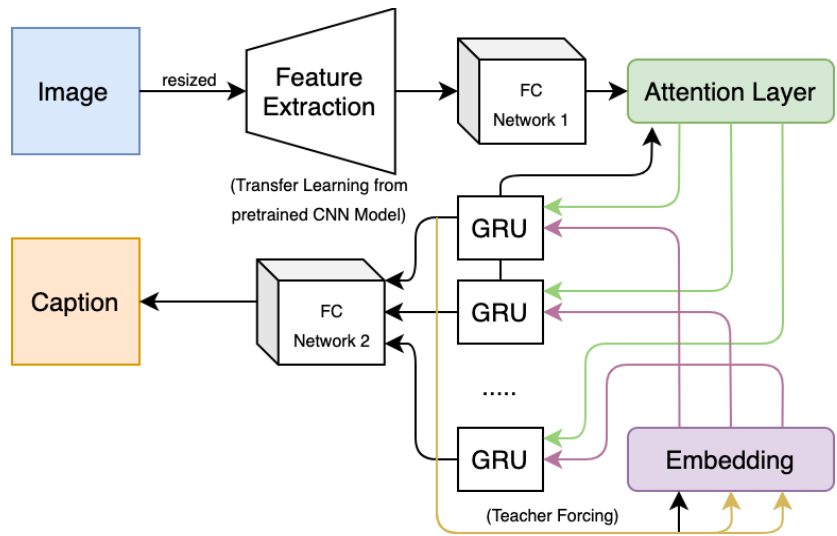

**Figure 3.** The image captioning network

ImageNet weight from the four architectures pre-trained model is used in the image feature extraction part. However, there are differences in the image input size for each architecture. EfficientNetB0 uses 224x224, EfficientNetB4 uses 380x380, InceptionV3 uses 299x299 and VGG16 uses 224x224. At the end of the feature extraction process, the image feature dimensions are also different for each architecture. EfficientNetB0 creates vectors in size of 49x1280, EfficientNetB4 in size of 121x1792, InceptionV3 in size of 64x2048 and VGG16 in size of 49x512. These feature vectors are then passed to a fully connected layer with 256 nodes. The resulting vector enters a loop as much as the maximum length of the training data caption through the Attention layer and then goes to the GRU-based decoder.

Within the Attention layer, we calculate the attention score and generate context vectors. The attention score is calculated from the activated summation of the value in the hidden state with the previous decoder layer. Softmax is used to finalize the score into probabilities values. The score will be multiplied with the image feature vector to be the context vector in the Attention layer. The context vector will be added with the embedding vector of words before entering the decoder. The concatenation context vector with a target word embedding vector is called the Teacher Forcing method. This technique helps learn the correct sequence or correct statistical property in front of the sequence, quickly. The decoder that we use is GRU with the initial initiation of Uniform Glorot numbers. The GRU layer has 512 hidden units which produce two different outputs, namely the word prediction matrix and the hidden state vector. The hidden state vector from decoder outputs will be used to train the next word. The prediction word vector is then passed to the fully connected layer with an output layer size equal to the existing vocabulary size, which is 291 units. In the training process, The predicted words will be used to calculate the loss with the target word embedding vector matrix. Finally, the gradient is calculated using an optimizer and runs the backpropagation to adjust the network parameter.

*3.3. Evaluation*

Since the aim of the image captioning model is to generate sentences that explain or express the given image, we evaluated the performance of the model's text generation using BLEU metrics. BLEU is a recognized metric for measuring the similarity of one hypothetical sentence to several reference sentences. We are given one hypothetical sentence and several reference sentences, returning a value between 0 and 1. A metric close to 1 means they are very similar [22]. In our evaluation, BLEU metrics take slices of the *n*-gram in the predicted caption and compare them to the ground truth caption data, and we use percentage normalization to represent the BLEU value. In our study, *n* is equal to 1.

## 4. Result and Discussion

EfficientNet is the state-of-the-art architecture used as a feature extractor in this research. We take the least and the highest architecture of EfficientNet, namely B0 and B4, in our experiment. We also take two other well-known models as a comparison (VGG16 and InceptionV3). Figure 4 shows the plot of the four training experiments we have done using 4 different models. The graphs in Figure 4 show a decline of loss as long as the epoch goes. The four architectures have similarities during the training. Around epoch 8 and 9 the four models started to suffer overfitting in about the same loss value. We choose the best checkpoint right about the intersection on evaluation lines, to be our image captioning model.

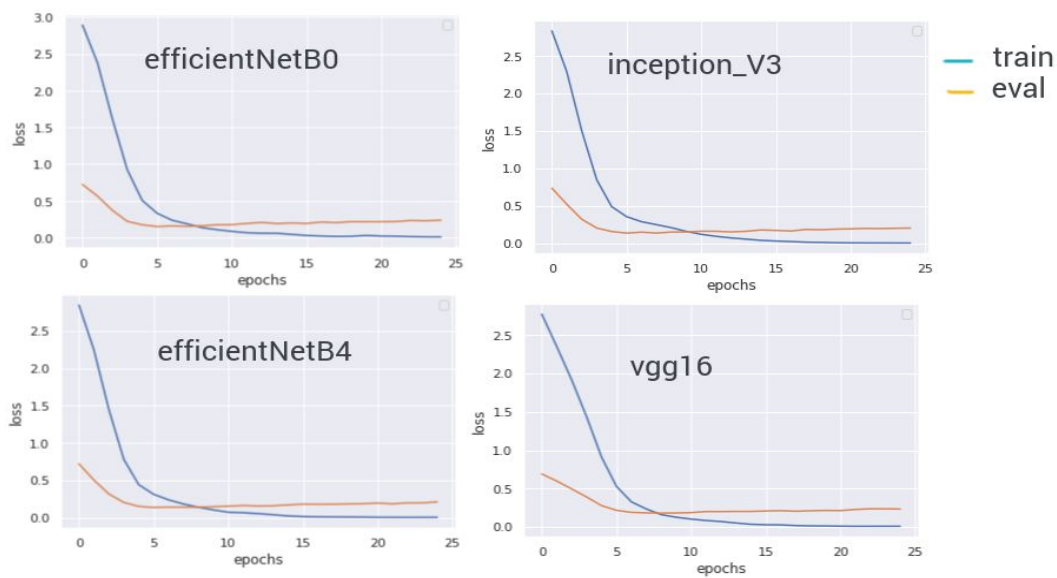

**Figure 4.** Training and evaluation plot of four different models

The four generated models are then evaluated using the BLEU metric. Table 1 shows the detailed average BLEU score of the four models. From the table, we could see that the EfficientNetB0 architecture is the best model for the context of our image captioning model using our collected dataset, with the average BLEU scores of 73.39 and 24.52 for training and validation respectively. The EfficientNetB4 model, which has a larger amount of parameters, has about a similar (slightly lower) performance to the B0 version. The result is interesting and fortunate to our local tourism domain. Inference resources can be saved to implement the smaller B0 model. Also, the EfficientNet model is proven to perform better than VGG16 and InceptionV3 in our local tourism domain image captioning. They have lower BLEU scores than the EfficientNet's.

**Table 1.** The Comparison of BLEU Metric Score of Four Different Models.

| Architecture | Average of the BLEU_train_score | Average of the BLEU_val_score |
| --- | --- | --- |
| EfficientNetB4 | 72.84 | 22.24 |
| EfficientNetB0 | **73.39** | **24.51** |
| VGG16 | 68.67 | 19.33 |
| InceptionV3 | 58.67 | 22.41 |

To show the inferencing result using our developed model which is based on EfficientNet, we take the captioning prediction result of three random validation images, as shown in Figure 5, Figure 6, and Figure 7. The first validation image, depicted in Figure 5, has a pretty high BLEU score in the predicted caption (71.42). The ground truth caption is *'menikmati ombak dan angin di pantai parangtritis'* (enjoying the waves and the wind on parangtritis beach) and the predicted caption is *'selalu ada dan angin di pantai parangtritis'* (always there and the wind at parangtritis beach). While the BLEU score is high, it still suffers from slight grammatical errors. This may be improved by feeding more grammatically structured training data. Another thing to highlight is that the predicted caption can point to specific tourism destinations. The second example, shown in Figure 6, is one of the interesting examples. It may not hold a high BLEU score, however, the predicted caption is sensible. The ground truth caption is *'serasa dapat terbang walaupun dengan bantuan ayunan'* (feels like I can fly even with the help of a swing) and the sensible predicted caption is *'banyak tumbuhan di bawahnya'* (many plants below). The third example that holds a pretty average BLEU score (42.85) also gives sensible predicted caption *'wisata merapi kaliadem private jeep tour yogyakarta'* (merapi tours kaliadem private jeep tour yogyakarta) which is close to the ground truth *'wisata jeep gunung merapi sleman'* (mount merapi jeep tour in sleman). Overall, the model has achieved a good sensible image captioning prediction in the domain of local tourism, especially in the region of Yogyakarta, Indonesia.

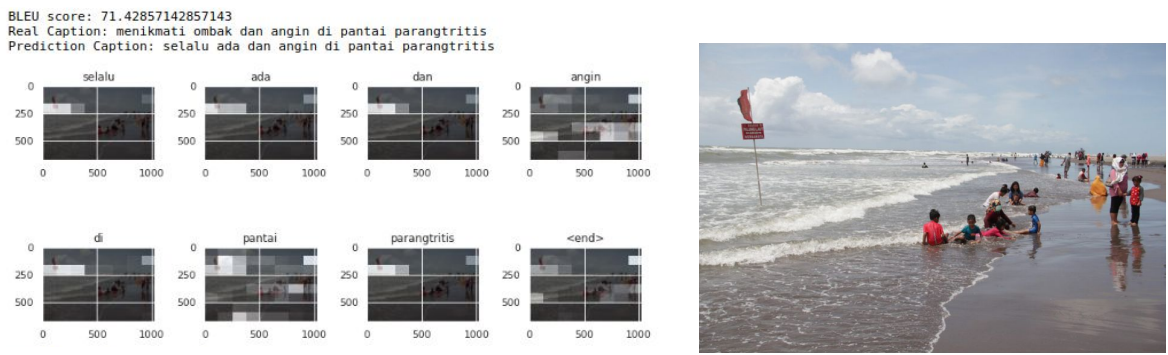

**Figure 5.** First image captioning prediction result

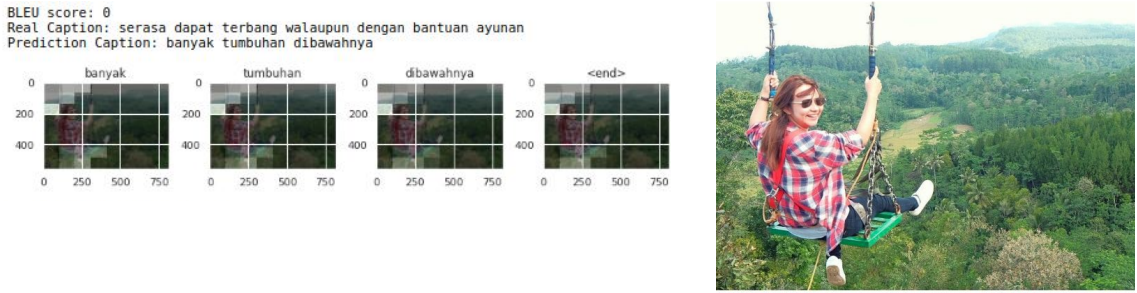

**Figure 6.** Second image captioning prediction result

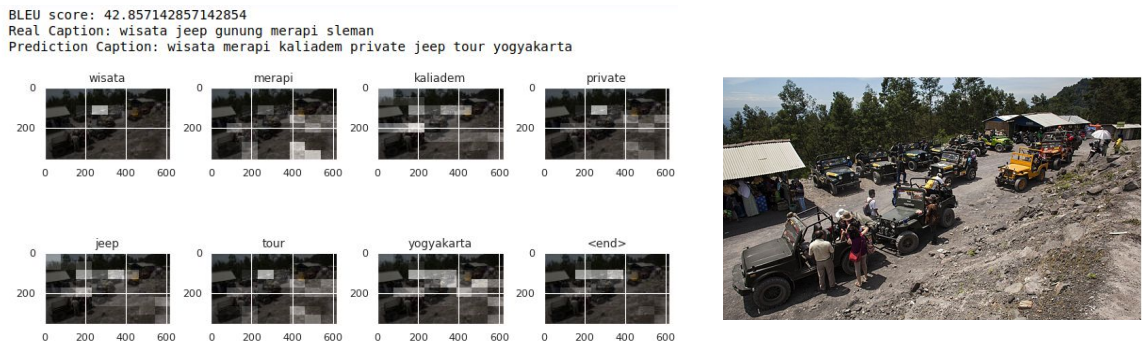

**Figure 7.** Third image captioning prediction result

## 5. Conclusion and Future Work

The information dissemination and engagement of smart local tourism systems to broader communities is crucial in tourism development. In this work, we have presented an attention-based image captioning on tourism images using EfficientNet to support a more humane engagement within the local tourism smart system. To evaluate the performance of the model, we compared the EfficientNet-based model to another well-known architecture model. Our experiment showed that our image captioning model using EfficientNetB0 and EfficientNetB4 architectures results in similar BLEU scores and got a higher score than another well-known trained model InceptionV3 and VGG16. However, EfficientNetB0 got a higher average train score and the average validation score than EfficientNetB4 with a score of 73.39 and 24.51 respectively. The captioning inference has shown that the EfficientNet-based image captioning model can produce a sensible yet detailed caption in the domain of local tourism. In the next development, we will create richer captions with more detailed objects or activities in the image to improve the suitability of the attention results in the captions. Besides, we will develop API wrapping for the image caption model that facilitates the integration of image captioning features with smart tourism systems, such as customer chatbot.